\documentclass[10pt,twocolumn,letterpaper]{article}

\usepackage{dicta}
\usepackage{times}
\usepackage{epsfig}
\usepackage{graphicx}
\usepackage{amsmath}
\usepackage{amssymb}
\usepackage{booktabs}
\usepackage{dblfloatfix}
\usepackage{multirow}

\usepackage[pagebackref=true,breaklinks=true,letterpaper=true,colorlinks,bookmarks=false]{hyperref}

\dictafinalcopy 

\def\dictaPaperID{57} 

\ifdictafinal\fi
\begin{document}

\title{MARC: Morphology-Aware Regression of Consensus for Cell Segmentation in Subcellular Spatial Transcriptomics}

\author{
Xinyu Shu$^{1,2}$ \quad
Andrew Zhang$^{1,2}$ \quad
Jean Yang$^{2,3,4,*}$ \quad
Jinman Kim$^{1,2,*}$ \\[3pt]
{\small $^{1}$School of Computer Science, The University of Sydney, Australia}\\
{\small $^{2}$Sydney Precision Data Science Centre, The University of Sydney, Australia}\\
{\small $^{3}$School of Mathematics and Statistics, The University of Sydney, Australia}\\
{\small $^{4}$Charles Perkins Centre, The University of Sydney, Australia}
}

\maketitle
\begin{abstract}
    Accurate cell segmentation remains a major bottleneck in subcellular spatial transcriptomics (SST), in which morphological images and spatially resolved RNA transcripts  are used to partition tissues into individual cellular instances. As segmentation serves as the foundation for constructing cell-level representations, boundary errors can lead to incorrect transcript assignments and compromise downstream analyses. However, reliable ground-truth boundaries are unavailable because they must be inferred from incomplete morphological and transcript signals. Furthermore, manual annotation of a large number of cells is time-consuming. Agreement among complementary segmentation methods provides a practical surrogate for identifying well-supported and ambiguous regions, but explicit consensus construction requires executing multiple computationally intensive pipelines. In this study, we propose MARC (Morphology-Aware Regression of Consensus), a framework that predicts a multi-method consensus-support map for SST segmentation. MARC is trained with leave-one-method-out consensus pseudo-targets and a Foreground-Union Consensus Loss that focuses supervision on candidate and consensus foreground. We evaluated MARC on 4,642 held-out tiles from Xenium kidney tissue, achieving a mean Dice score of 0.90, a mean intersection-over-union of 0.82, and a mean cell-level Spearman correlation of 0.79 against explicitly computed cross-method consensus maps. We demonstrate that the predicted consensus maps localise weakly supported regions while preserving consensus-based rankings and identifying low-consensus cells for manual review. These results show that MARC closely approximates explicit cross-method consensus without multi-method inference and therefore has the potential to facilitate robust, consensus-aware evaluation of cell segmentation in large-scale SST studies.
\end{abstract}

\section{Introduction}
Subcellular spatial transcriptomics (SST) combines high-resolution tissue imaging with spatially resolved RNA transcripts to quantify cellular organisation within intact tissue~\cite{chen2015multiplexed,eng2019seqfish,rodriques2019slideseq,stahl2016spatial}. Cell segmentation converts these combined imaging and transcript measurements into cell-level profiles by assigning transcripts to individual cells. Large SST tissue regions contain many densely packed cells, while weak morphological contrast, sparse or displaced transcripts, and overlapping neighbouring cells make their boundaries difficult to determine~\cite{ishaque2026challenge}. As SST datasets scale, segmentation-derived cell-level measurements increasingly underpin downstream biological discovery, making segmentation quality control an essential component of computational tissue analysis.

\begin{figure*}[t]
    \centering
    \includegraphics[width=\textwidth]{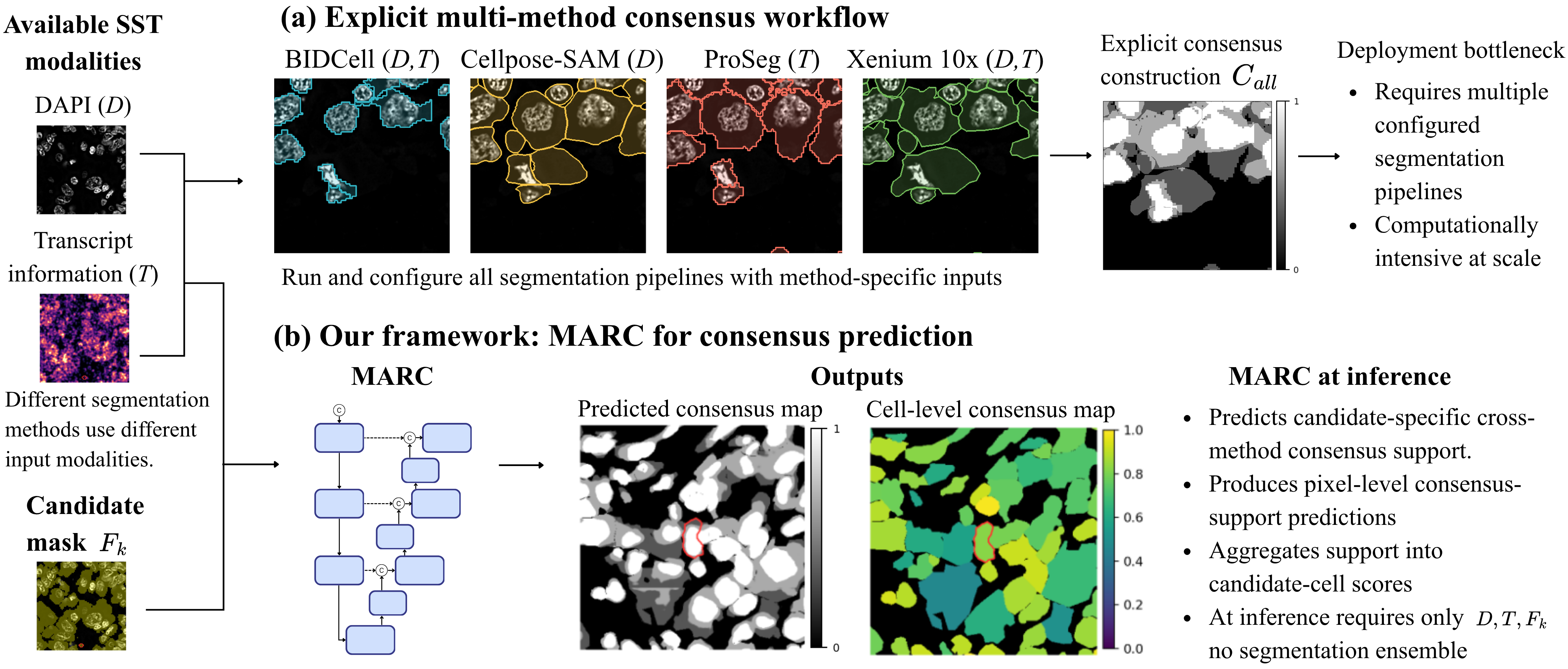}
    \caption{Overview of explicit multi-method consensus construction and MARC. (a) Explicit consensus averages outputs from multiple segmentation pipelines using method-specific SST inputs, making consensus construction computationally intensive. (b) MARC predicts pixel-level consensus support from a candidate mask $F_k$, DAPI morphology $D$, and transcript information $T$, and aggregates the prediction into cell-level scores. MARC requires no segmentation ensemble at inference. The same candidate cell is outlined in red.}
    \label{fig:segmentation_problem}
\end{figure*}

Current SST segmentation methods use morphological images and transcript measurements in different ways, as illustrated in Figure~\ref{fig:segmentation_problem}(a). Methods such as Cellpose-SAM, BIDCell, ProSeg, and Xenium rely on different inputs and assumptions and can therefore produce conflicting boundaries for the same tissue region~\cite{tenx2025cellsegmentation,fu2024bidcell,jones2025proseg,pachitariu2025cellposesam}. This disagreement is particularly pronounced in densely packed tissue and regions with weak or ambiguous signals. Conventional segmentation quality-assessment approaches were not developed specifically for SST cell segmentation. Supervised approaches require expert reference masks, which are time-consuming to produce because SST datasets contain large numbers of cells, while uncertainty-based approaches require access to internal predictions from the model that generated the segmentation~\cite{gal2016dropout,kendall2017uncertainties,specktorfadida2025segqc}. Consequently, these approaches do not readily support the evaluation of candidate masks produced by different SST segmentation pipelines when expert annotations are unavailable. 

Because SST cell boundaries are inherently ambiguous and expert annotations are prohibitively time-consuming to produce at scale, agreement among complementary segmentation methods provides a practical surrogate for identifying well-supported and ambiguous cellular regions. Consensus estimation has previously been used to infer latent segmentation support from multiple imperfect observations rather than treating any individual segmentation as ground truth~\cite{warfield2004staple}. In SST, morphology-based and transcript-aware segmentation methods provide complementary evidence because their errors arise from different inputs and assumptions. However, constructing an explicit cross-method consensus requires configuring and running all contributing pipelines with their method-specific SST inputs, as illustrated in Figure~\ref{fig:segmentation_problem}(a). For tissue regions containing large numbers of cells, this process is computationally intensive and difficult to scale, limiting its use for routine quality control.

We propose MARC (Morphology-Aware Regression of Consensus), a framework that predicts consensus support for a candidate SST segmentation. As illustrated in Figure~\ref{fig:segmentation_problem}(b), MARC is trained using leave-one-method-out consensus pseudo-targets. At inference, it requires only a candidate mask, aligned DAPI-derived morphology, and transcript-density information. It produces a dense consensus-support map that is aggregated into cell-level quality-control scores.

Our contributions are threefold. First, we formulate quality control for candidate SST segmentations as a dense consensus-regression problem that approximates leave-one-method-out agreement without constructing an explicit consensus at inference. Second, we introduce MARC, which combines candidate-mask geometry with aligned morphological and transcript signals using consensus-derived pseudo-supervision. Its Foreground-Union Consensus Loss (FUCL) focuses training on candidate and consensus foreground, including disagreement regions. Third, we evaluate MARC at both the pixel and cell levels across four candidate segmentation methods on a spatially held-out region of a Xenium renal cell carcinoma dataset~\cite{tenx2025rccdataset}.

\section{Related work}
\subsection{Cell segmentation in SST}

Classical cell segmentation frameworks such as CellProfiler provide flexible rule-based workflows for microscopy, while deep learning methods learn segmentation directly from annotated data \cite{mcquin2018cellprofiler}. U-Net established the encoder--decoder architecture as a standard backbone for biomedical segmentation \cite{ronneberger2015unet}. Subsequent methods, including Mask R-CNN, HoVer-Net, and Cellpose-SAM, extended these advances to instance segmentation, nuclear analysis, and generalist cell segmentation across diverse imaging modalities \cite{graham2019hovernet,he2017maskrcnn,pachitariu2025cellposesam}. Together, these methods have substantially improved segmentation accuracy across a wide range of biological imaging applications.

SST introduces additional challenges because segmentation determines how transcripts are assigned to individual cells, directly affecting downstream analyses such as cell-type annotation, spatial-interaction modelling, and tissue organisation. Consequently, visually plausible boundaries do not necessarily correspond to biologically meaningful transcript assignments, particularly in regions with weak morphology, sparse transcripts, or densely packed cells \cite{ishaque2026challenge}.

Transcript-informed methods address this problem by incorporating molecular measurements. Baysor and ProSeg infer cell membership from transcript locations and gene identity, while graph-based methods group 
transcripts using spatial relationships \cite{andersson2023signedgraph,jones2025proseg,petukhov2022baysor}. BIDCell uses biologically informed self-supervised learning to integrate morphological and transcript information for SST segmentation \cite{fu2024bidcell}. These methods use different assumptions and input modalities, resulting in complementary but potentially inconsistent candidate segmentation masks. Most existing work focuses on generating improved masks rather than evaluating the reliability of an externally generated candidate mask.

\subsection{Segmentation quality estimation}
Segmentation quality is commonly measured against manually annotated masks. In SST, however, dense manual annotations are expensive, and cell boundaries may be biologically ambiguous~\cite{ishaque2026challenge}. Learning-based quality-control methods instead seek to estimate segmentation quality directly from images and candidate masks.

SegQC predicts segmentation-error maps and quality metrics from expert-corrected medical image segmentations~\cite{specktorfadida2025segqc}, while Bayesian uncertainty methods estimate uncertainty from a model's own predictions~\cite{gal2016dropout,kendall2017uncertainties}. Neither approach was developed specifically for SST. SegQC relies on expert-corrected masks, whereas Bayesian uncertainty methods require access to the model that generated the segmentation. MARC instead learns from consensus-derived pseudo-targets rather than manual ground truth or internal model uncertainty.

\subsection{Consensus methods for segmentation}

Consensus methods combine multiple imperfect segmentations when no single output can be assumed correct. STAPLE estimates a probabilistic latent segmentation while accounting for the performance of individual raters or methods, whereas majority voting provides a simpler agreement estimate~\cite{warfield2004staple}.

However, computing conventional consensus requires running every contributing pipeline with its method-specific SST inputs, making consensus construction computationally intensive at scale. MARC instead learns to predict cross-method consensus from a candidate mask with aligned DAPI and transcript density, avoiding multi-method inference.
\section{Methods}
\begin{figure*}[t]
    \centering
    \includegraphics[width=\textwidth]{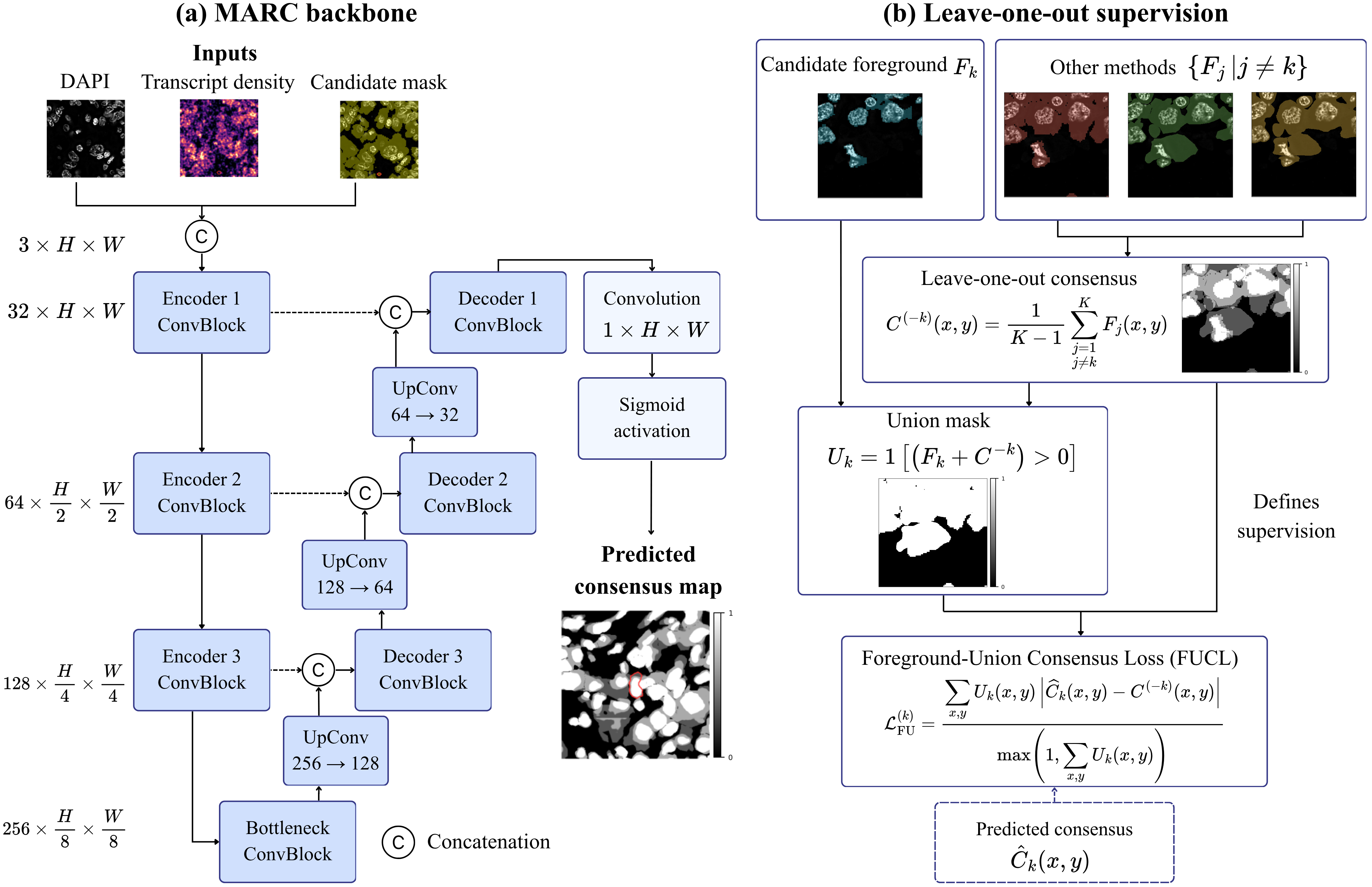}
    \caption{Architecture of MARC for predicting pixel-level consensus support from a candidate mask, DAPI morphology, and transcript density.}
    \label{fig:marc_architecture}
\end{figure*}

\subsection{Problem formulation}
Let \(M_k\in\{0,1,\ldots\}^{H\times W}\) denote the labelled mask produced by candidate method \(k\), where positive values identify cell instances and zero denotes background. Its binary foreground map is

\begin{equation}
F_k(x,y) =
\begin{cases}
1, & M_k(x,y) > 0,\\
0, & M_k(x,y) = 0.
\end{cases}
\label{eq:binary_foreground}
\end{equation}

Let \(D \in \mathbb{R}^{H \times W}\) denote the aligned DAPI morphology image and \(T \in \mathbb{R}^{H \times W}\) the transcript-density image. For each \(k\), MARC concatenates \(D\), \(T\), and \(F_k\) to form a three-channel input:

\begin{equation}
\mathbf{X}_k
=
\operatorname{concat}
\left(D,\,T,\,F_k\right)
\in \mathbb{R}^{3 \times H \times W }.
\label{eq:model_input}
\end{equation}

Given \(\mathbf{X}_k\), the model \(f_{\theta}\), parameterised by \(\theta\), predicts a continuous consensus-support map:

\begin{equation}
\widehat{C}_k
=
f_{\theta}\!\left(\mathbf{X}_k\right),
\qquad
\widehat{C}_k \in [0,1]^{H \times W}.
\label{eq:consensus_prediction}
\end{equation}

Values of \(\widehat{C}_k\) near one indicate strong consensus agreement, while values near zero indicate weak agreement. The labelled mask \(M_k\) is retained for cell-level aggregation.

\subsection{MARC architecture}
MARC uses a four-level U-Net-style encoder--decoder~\cite{ronneberger2015unet}. Each encoder block contains two \(3\times3\) convolution--batch-normalisation--ReLU layers, with channel widths of 32, 64, 128, and 256, and \(2\times2\) max pooling between levels. The decoder uses \(2\times2\) transposed convolutions, skip concatenation with the corresponding encoder features, and the same convolutional blocks. A final \(1\times1\) convolution produces a single-channel logit map \(Z_k\), followed by a sigmoid:

\begin{equation}
\widehat{C}_k
=
f_{\theta}\!\left(\mathbf{X}_k\right)
=
\sigma\!\left(Z_k\right)
=
\frac{1}{1+\exp\!\left(-Z_k\right)}.
\label{eq:sigmoid_output}
\end{equation}

\subsection{Leave-one-method-out consensus supervision}

Manual boundary annotations are not available for the full tissue. We therefore construct pseudo-supervision from leave-one-method-out consensus. Let \(K\) denote the total number of candidate segmentation methods. For candidate method \(k\), the consensus target is calculated as the mean binary foreground prediction of the remaining \(K-1\) methods:

\begin{equation}
C^{(-k)}(x,y)
=
\frac{1}{K-1}
\sum_{\substack{j=1\\j\neq k}}^{K}
F_j(x,y).
\label{eq:loo_consensus}
\end{equation}
The resulting map \(C^{(-k)}\in [0,1]^{H \times W}\) is used as the pseudo-target during training and as the reference during evaluation. In our experiments, \(K=4\); therefore, each target is the mean foreground consensus of the other three segmentation methods. Excluding the candidate method from its own target prevents the model from receiving trivial self-agreement as supervision.

\begin{table*}[!b]
\caption{Performance of MARC predictions against explicit consensus targets. The best result for each metric is shown in bold.}
\label{tab:consensus_prediction}
\centering
\small
\setlength{\tabcolsep}{5pt}
\begin{tabular}{@{}lcccccc@{}}
\toprule
Candidate input
& Foreground-union L1 \(\downarrow\)
& Dice \(\uparrow\)
& IoU \(\uparrow\)
& Sensitivity \(\uparrow\)
& Precision \(\uparrow\)
& Cell Spearman correlation \(\rho\) \(\uparrow\) \\
\midrule
Cellpose-SAM
& 0.0832 & 0.9007 & 0.8193 & 0.9020 & \textbf{0.8994} & 0.8172 \\
BIDCell
& 0.1060 & 0.9029 & 0.8230 & \textbf{0.9298} & 0.8774 & 0.6251 \\
ProSeg
& 0.1240 & 0.8844 & 0.7927 & 0.9073 & 0.8626 & \textbf{0.8802} \\
Xenium 10x
& \textbf{0.0793} & \textbf{0.9127} & \textbf{0.8394}
& 0.9283 & 0.8977 & 0.8396 \\
\midrule
\textbf{Avg. over methods}
& 0.0981 & 0.9002 & 0.8186 & 0.9168 & 0.8843 & 0.7905 \\
\bottomrule
\end{tabular}
\end{table*}

\subsection{Foreground-Union Consensus Loss (FUCL)}

We define a foreground-union mask that includes every pixel belonging to either the candidate foreground or the leave-one-method-out consensus foreground:

\begin{equation}
U_k(x,y)
=
\begin{cases}
1, & F_k(x,y)=1
     \ \lor\ 
     C^{(-k)}(x,y)>0,\\
0, & \text{otherwise}.
\end{cases}
\label{eq:foreground_union}
\end{equation}

MARC is trained using FUCL, defined as the mean absolute error within this union:

\begin{equation}
\mathcal{L}_{\mathrm{FU}}^{(k)}
=
\frac{
\displaystyle
\sum_{x,y}
U_k(x,y)
\left|
\widehat{C}_k(x,y)-C^{(-k)}(x,y)
\right|
}{
\displaystyle
\max\!\left(
1,\sum_{x,y}U_k(x,y)
\right)
}.
\label{eq:foreground_union_loss}
\end{equation}

Clamping the denominator avoids division by zero, while restricting the loss to \(U_k\) focuses learning on foreground and disagreement regions.

\subsection{Cell-level consensus aggregation}

The dense map \(\widehat{C}_k\) is aggregated into cell-level scores using the original labelled mask \(M_k\). For each candidate cell \(c\), we average the predicted consensus support over the pixels assigned to that cell:

\begin{equation}
S_{k,c}
=
\frac{
\displaystyle
\sum_{(x,y):\,M_k(x,y)=c}
\widehat{C}_k(x,y)
}{
\displaystyle
\left|
\left\{(x,y):M_k(x,y)=c\right\}
\right|
}.
\label{eq:cell_score}
\end{equation}
The denominator denotes the number of pixels assigned to candidate cell \(c\). This produces one consensus-support score for each candidate cell. Target cell-level scores are computed analogously by replacing \(\widehat{C}_k\) with \(C^{(-k)}\). Cell-level Spearman correlation is then calculated between the predicted and target scores across candidate cells. These scores can be used to rank cells for manual review, filtering, or downstream quality-control analysis.
\section{Results and Discussion}

\subsection{Dataset and setup}
We evaluated MARC on the public 10x Genomics Xenium FFPE human renal
cell carcinoma dataset~\cite{tenx2025rccdataset}. Candidate masks were
generated using Cellpose-SAM, BIDCell, ProSeg, and Xenium onboard
segmentation~\cite{tenx2025cellsegmentation,fu2024bidcell,jones2025proseg, pachitariu2025cellposesam}. 

All inputs were aligned to a \(0.2125~\mu\mathrm{m}/\mathrm{pixel}\) grid. Spatial splitting and \(256\times256\) tiling yielded 4,642 training, 2,321 validation, and 4,642 test tiles. Each candidate-method example paired one spatial tile with one candidate segmentation method. Pairs containing less than \(1\%\) candidate foreground were excluded, yielding 17,259 training, 9,058 validation, and 18,443 test examples.

MARC was trained using AdamW~\cite{loshchilov2019decoupled}
with a batch size of 12, a learning rate of
\(3 \times 10^{-4}\), and the FUCL; the best validation checkpoint was selected. The primary supervision target was the explicit leave-one-method-out mean consensus defined in Eq.~\eqref{eq:loo_consensus}, obtained by averaging the binary foreground maps of the other three methods. 

Foreground-union L1 and cell-level Spearman correlation were calculated from continuous support values, whereas Dice, IoU, sensitivity, and precision were calculated after thresholding predictions and targets at \(0.5\), which corresponds to foreground support from a majority of the other three methods.

\begin{figure*}[t]
    \centering
    \includegraphics[width=\textwidth]{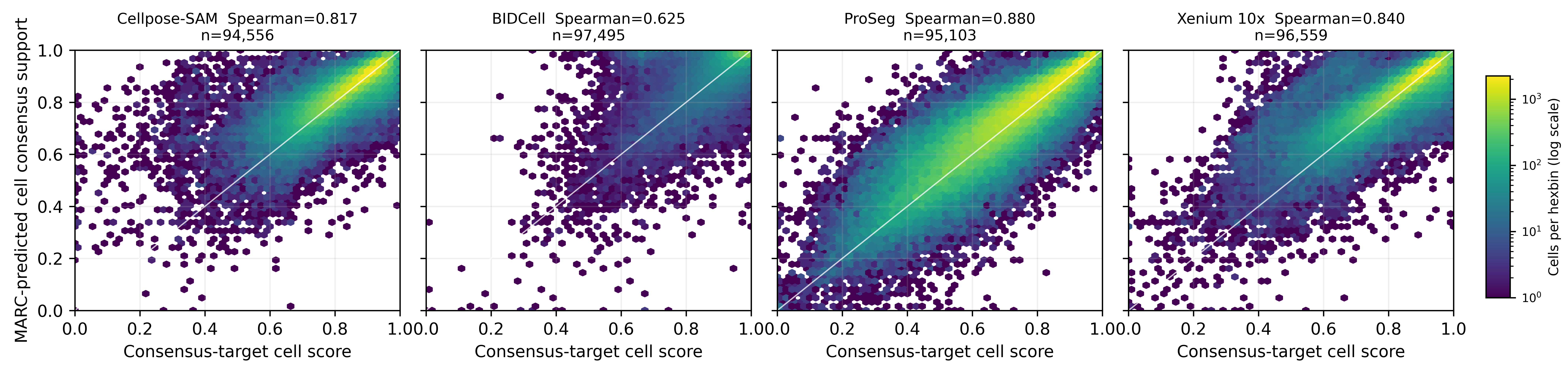}
    \caption{Predicted versus target cell-level consensus-support scores for
each candidate method on the held-out test region.}
    \label{fig:cell_score_correlation}
\end{figure*}

\begin{table*}[!b]
\caption{Input-channel ablation evaluated against explicit consensus targets.}
\label{tab:input_ablation}
\centering
\small
\renewcommand{\arraystretch}{1.05}
\setlength{\tabcolsep}{4pt}
\begin{tabular}{@{}llccccc@{}}
\toprule
Configuration & Method
& Dice \(\uparrow\)
& IoU \(\uparrow\)
& Sensitivity \(\uparrow\)
& Precision \(\uparrow\)
& Cell Spearman correlation \(\rho\) \(\uparrow\) \\
\midrule

\multirow{5}{*}{\shortstack[l]{Candidate-copy\\baseline}}
& BIDCell      & 0.6990 & 0.5372 & 0.5474 & 0.9665 & --\(^{\ast}\) \\
& Cellpose-SAM & 0.8878 & 0.7982 & 0.8794 & 0.8963 & --\(^{\ast}\) \\
& ProSeg       & 0.8028 & 0.6705 & 0.9608 & 0.6894 & --\(^{\ast}\) \\
& Xenium 10x   & 0.8863 & 0.7958 & 0.9070 & 0.8665 & --\(^{\ast}\) \\
& \textbf{Avg. over methods}
               & 0.8189 & 0.7004 & 0.8237 & 0.8547 & --\(^{\ast}\) \\
\midrule

\multirow{5}{*}{\shortstack[l]{MARC: candidate\\mask only}}
& BIDCell      & 0.8364 & 0.7189 & 0.8031 & 0.8727 & 0.2493 \\
& Cellpose-SAM & 0.8893 & 0.8007 & 0.8946 & 0.8842 & 0.4781 \\
& ProSeg       & 0.8089 & 0.6792 & 0.9094 & 0.7284 & 0.5019 \\
& Xenium 10x   & 0.8913 & 0.8040 & 0.9234 & 0.8614 & 0.4617 \\
& \textbf{Avg. over methods}
               & 0.8565 & 0.7507 & 0.8826 & 0.8367 & 0.4228 \\
\midrule

\multirow{5}{*}{\shortstack[l]{MARC: candidate\\mask + DAPI}}
& BIDCell      & 0.8718 & 0.7728 & 0.8648 & 0.8790 & 0.5562 \\
& Cellpose-SAM & 0.8933 & 0.8072 & 0.8870 & 0.8998 & 0.7824 \\
& ProSeg       & 0.8755 & 0.7786 & 0.8813 & 0.8698 & 0.8743 \\
& Xenium 10x   & 0.9058 & 0.8278 & 0.9132 & 0.8985 & 0.8129 \\
& \textbf{Avg. over methods}
               & 0.8866 & 0.7966 & 0.8866 & \textbf{0.8868} & 0.7564 \\
\midrule

\multirow{5}{*}{\shortstack[l]{MARC: candidate mask\\+ DAPI + transcripts}}
& BIDCell      & 0.9029 & 0.8230 & 0.9298 & 0.8774 & 0.6251 \\
& Cellpose-SAM & 0.9007 & 0.8193 & 0.9020 & 0.8994 & 0.8172 \\
& ProSeg       & 0.8844 & 0.7927 & 0.9073 & 0.8626 & 0.8802 \\
& Xenium 10x   & 0.9127 & 0.8394 & 0.9283 & 0.8977 & 0.8396 \\
& \textbf{Avg. over methods}
               & \textbf{0.9002} & \textbf{0.8186}
               & \textbf{0.9168} & 0.8843 & \textbf{0.7905} \\
\bottomrule

\multicolumn{7}{@{}l}{\footnotesize
\(^{\ast}\)Undefined because the candidate-copy baseline produces
constant cell-level scores.}
\end{tabular}
\end{table*}

\subsection{Consensus prediction}
Table~\ref{tab:consensus_prediction} evaluates how closely MARC’s predicted consensus-support maps, \(\widehat{C}_k\), reproduce the explicitly computed leave-one-method-out mean-consensus targets, \(C^{(-k)}\), defined in Eq.~\eqref{eq:loo_consensus}. In each row, the named candidate method supplies the input mask to MARC and is excluded from the explicit consensus target, which is constructed from the remaining three methods. For example, in the Cellpose-SAM row, MARC predicts consensus support for the Cellpose-SAM candidate mask, while the explicit target is calculated from BIDCell, ProSeg, and Xenium. The Dice score of \(0.9007\) therefore indicates strong overlap between MARC's prediction and the explicit consensus target. 

Across all candidate methods, MARC achieved mean foreground-union L1, Dice, and IoU values of \(0.0981\), \(0.9002\), and \(0.8186\), respectively. The low L1 and high overlap metrics indicate that MARC closely reproduces the explicit consensus at the pixel level. Cell-level Spearman correlations ranged from \(0.6251\) to \(0.8802\), with a mean of \(0.7905\), indicating that MARC also preserves the ordering of candidate cells from low to high consensus support.

Figure~\ref{fig:cell_score_correlation} shows the relationship between predicted and explicit-target cell-level consensus-support scores for each candidate method. The proposed aggregation converts the dense prediction into meaningful candidate-cell scores that can be used to prioritise low-consensus cells for review.

\subsection{Ablation study}

\begin{figure*}[t]
    \centering
    \includegraphics[width=\textwidth]{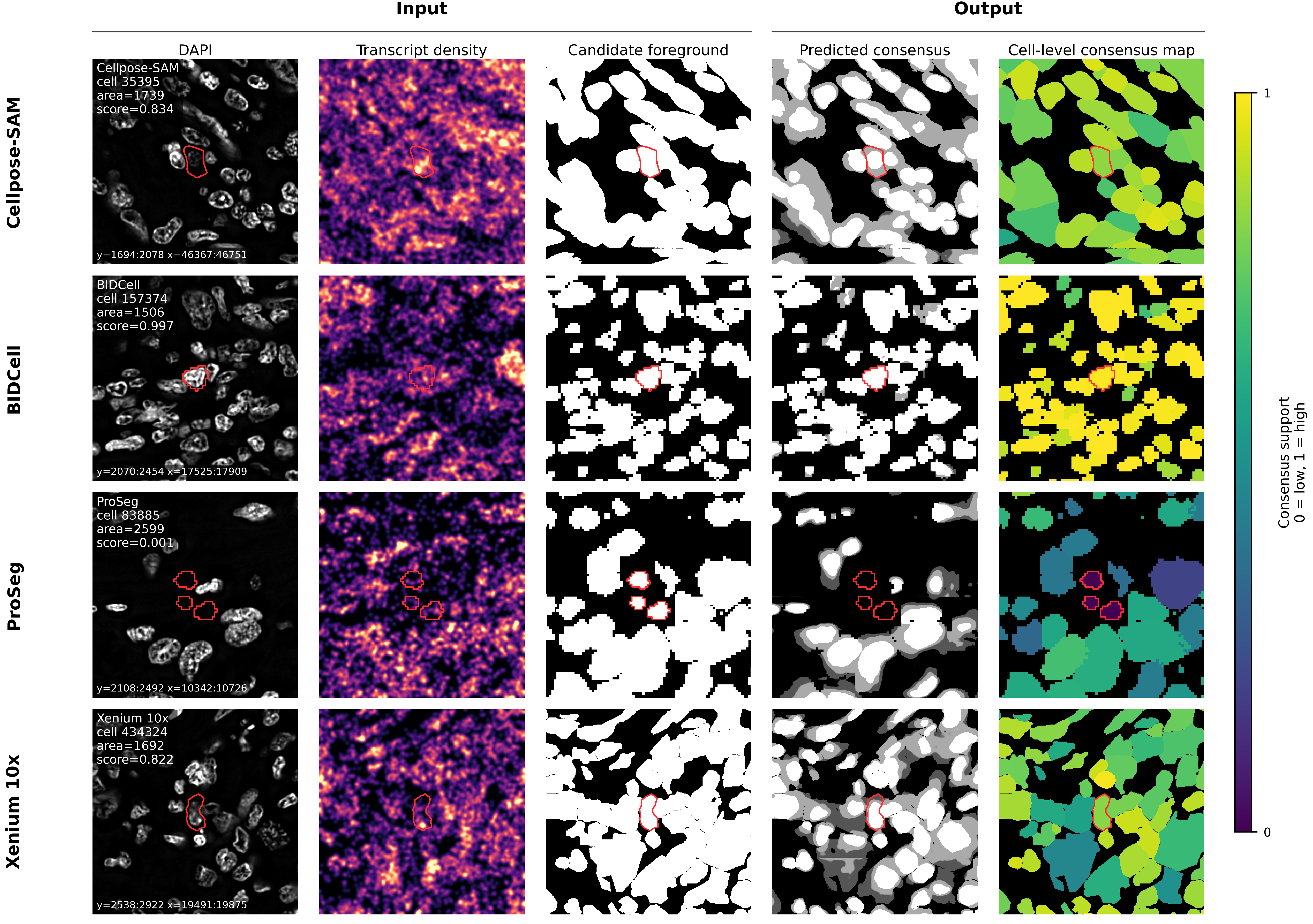}
    \caption{Qualitative consensus-support predictions on the held-out test region. Rows show representative candidate cells from each segmentation method, while columns show the DAPI morphology, transcript-density map, candidate foreground mask, predicted pixel-level consensus, and predicted cell-level consensus map. Red outlines delineate the representative candidate cell in each row. Legends in the first column report the segmentation method, cell identifier, area in pixels, predicted cell-level consensus score, and crop coordinates. Brighter values in the output maps indicate stronger predicted cross-method consensus support.}
    \label{fig:qualitative_consensus}
\end{figure*}

\noindent\textbf{Target ablation.}
To assess whether MARC depends on the consensus construction, we trained two full-input MARC models that differed only in their supervision targets. For each candidate method \(k\), the primary model used the arithmetic mean of the other three foreground masks, whereas the alternative model used a STAPLE fusion of the same three masks~\cite{warfield2004staple}. STAPLE uses expectation--maximisation to jointly estimate a probabilistic latent foreground map and the sensitivity and specificity of each contributing method, thereby weighting the input masks according to their estimated performance. Both models used the candidate mask, DAPI morphology, and transcript-density map as inputs, with identical architecture, data splits, loss, and optimisation settings.

The models were evaluated on the same held-out candidate examples against their corresponding leave-one-method-out targets. Mean-consensus and STAPLE supervision produced similar average Dice scores (\(0.9002\) versus \(0.8992\)) and IoU values (\(0.8186\) versus \(0.8169\)). Mean consensus achieved higher sensitivity (\(0.9168\) versus \(0.8860\)) and cell-level Spearman correlation (\(0.7905\) versus \(0.7491\)), whereas STAPLE achieved higher precision (\(0.9128\) versus \(0.8843\)). Thus, the two target formulations yielded comparable pixel-level overlap against their respective targets but different trade-offs: mean-consensus supervision achieved higher sensitivity and better preserved cell-level rankings, whereas STAPLE supervision achieved higher precision. Because each model was evaluated against a different target, this comparison does not establish that one target is more accurate. We retained mean consensus because it directly represents the fraction of supporting methods and is used in the primary evaluation.

\noindent\textbf{Input ablation.}
Table~\ref{tab:input_ablation} compares a direct candidate-copy baseline with three separately trained MARC variants: candidate mask only, candidate mask with DAPI, and candidate mask with DAPI and transcript density. All MARC variants used the same leave-one-method-out mean-consensus supervision and were evaluated against the same explicit mean-consensus targets; only their input channels differed.

The candidate-copy baseline directly treats the candidate foreground mask as its consensus prediction and achieved an average Dice score of \(0.8189\). Because every foreground pixel within a candidate cell is assigned a value of one, its aggregated cell-level scores are constant and cannot rank cells by consensus support.

The candidate-mask-only MARC model increased the average Dice score to 0.8565 and achieved a cell-level Spearman correlation of 0.4228. Adding DAPI morphology further improved the average Dice score to 0.8866 and the cell-level Spearman correlation to 0.7564. Adding transcript density to the candidate-mask-and-DAPI model produced the best overall performance, with an average Dice score of 0.9002, IoU of 0.8186, and cell-level Spearman correlation of 0.7905. The substantial increase in cell-level correlation after adding DAPI indicates that morphology provides the main additional signal for ranking candidate cells. Transcript density contributes complementary molecular information and further improves both pixel-level consensus prediction and cell-level ranking. These results indicate that MARC uses contextual evidence beyond candidate-mask geometry.

\subsection{Qualitative consensus-support analysis}

Figure~\ref{fig:qualitative_consensus} shows representative examples from the held-out test region. For each candidate method, MARC predictions are presented alongside the DAPI morphology, transcript-density map, and candidate foreground mask. 

The cell-level map assigns each candidate cell the mean predicted support across its foreground pixels. High predicted consensus was generally observed in candidate regions aligned with nuclear morphology, transcript density, and cross-method foreground support. Lower consensus was observed where candidate masks extended into the background, disagreed with local morphological or molecular evidence, or represented method-specific detections with weak support from the other segmentation methods. These examples show that MARC localises shared and weakly supported regions and produces spatially varying predictions.

\subsection{Limitations}
A key limitation of this study is that consensus maps remain a surrogate rather than the ground truth, i.e., failure modes across segmentation methods may be reinforced. Nonetheless, we suggest that such a surrogate is useful because it captures agreement across multiple segmentation methods, providing a pragmatic and scalable alternative to manual annotation while reducing reliance on any single method's biases. Another limitation is that the MARC was evaluated on a single Xenium renal cell carcinoma dataset, which limits generalisability. The candidate methods also share inputs from the same Xenium assay and are therefore not fully independent. Future work will evaluate additional tissues, platforms, and segmentation methods, incorporate orthogonal membrane or protein markers for validation, and determine whether consensus-guided filtering improves downstream analyses.

\section{Conclusion}
\label{sec:conclusion}
In this study, we proposed a new framework termed MARC (Morphology-Aware Regression of Consensus) for predicting cross-method consensus support for candidate SST segmentations. MARC learns from leave-one-method-out consensus pseudo-targets and, at inference, requires only a single candidate mask together with aligned DAPI morphology and transcript-density information. Our results demonstrated that MARC closely approximated explicitly computed consensus at both pixel and cell levels. Input ablations further showed that morphological and molecular context provided complementary information beyond candidate-mask geometry. By producing spatially resolved consensus-support maps, MARC provides a practical approach for large-scale SST analysis.

{\small
\bibliographystyle{ieee}
\bibliography{egbib}
}

\end{document}